\documentclass[a4paper]{article}
\usepackage{Odyssey2024}
\usepackage{epsfig,amssymb,amsmath}
\usepackage{booktabs}
\usepackage{multirow}
\usepackage{graphicx}
\usepackage{subcaption}
\usepackage{float}
\usepackage{numprint}
\nprounddigits{3}
\npdecimalsign{.}
\usepackage{pgfplots}
\usepackage{tikz}
\pgfplotsset{compat=1.17}
\ninept

\usepackage[acronym]{glossaries}
\newacronym{SVM}{SVM}{Support Vector Machine}
\newacronym{NLP}{NLP}{Natural Language Processing}
\newacronym{SSL}{SSL}{Self-Supervised Learning}
\newacronym{SER}{SER}{Speech Emotion Recognition}

\title{MSP-Podcast SER Challenge 2024: L'antenne du Ventoux \\ Multimodal Self-Supervised Learning for Speech Emotion Recognition}

\makeatletter
\def\name#1{\gdef\@name{#1\\}}
\makeatother
\name{{\em Jarod Duret, Mickael Rouvier, Yannick Est{\`e}ve}}

\address{LIA  \\
 Avignon Universite, France \\
{\small \tt first.last@univ-avignon.fr} }
\begin{document}
\maketitle

\begin{abstract}
In this work, we detail our submission to the 2024 edition of the MSP-Podcast Speech Emotion Recognition (SER) Challenge. This challenge is divided into two distinct tasks: Categorical Emotion Recognition and Emotional Attribute Prediction. We concentrated our efforts on Task 1, which involves the categorical classification of eight emotional states using data from the MSP-Podcast dataset. Our approach employs an ensemble of models, each trained independently and then fused at the score level using a Support Vector Machine (SVM) classifier. The models were trained using various strategies, including Self-Supervised Learning (SSL) fine-tuning across different modalities: speech alone, text alone, and a combined speech and text approach. This joint training methodology aims to enhance the system's ability to accurately classify emotional states. This joint training methodology aims to enhance the system's ability to accurately classify emotional states. Thus, the system obtained F1-macro of 0.35\% on development set.
\end{abstract}



%
\section{Introduction}
\gls{SER} represents a challenging area of research.   
The complexity arises from the nuanced, subjective nature of emotional expression in speech and the challenge of extracting effective feature representations.  
Despite these difficulties, understanding emotions is crucial for enhancing human-computer interaction, as emotions significantly influence reasoning, decision-making, and social interactions.  
In the realm of speech and text, emotions, while subjective, are essential to convey meaning and intent.   
In recent years, advances in deep learning have contributed to notable improvements in the performance of emotion recognition systems by leveraging highly effective features extracted from deep neural networks.  \newline
In the pursuit of advancing \gls{SER} capabilities, the MSP-Podcast \gls{SER} Challenge 2024\cite{Goncalves_2024} provides a fertile ground for exploring novel methodologies and techniques for emotion recognition from naturalistic speech data.
Our focus lies primarily on Task~1: Categorical Emotion Recognition, which involves classifying speech segments into eight specified emotional states: Anger (A), Happiness (H), Sadness (S), Fear (F), Surprise (U), Contempt (C), Disgust (D), and Neutral (N).

This paper details our submission to the 2024 edition of the MSP-Podcast \gls{SER} Challenge. Our approach employs an ensemble of models, each trained independently and then fused at the score level using a Support Vector Machine (SVM) classifier. The models were trained using various strategies, including Self-Supervised Learning (SSL) fine-tuning across different modalities: speech alone, text alone, and a combined speech and text approach. This joint training methodology aims to enhance the system's ability to accurately classify emotional states. 


The paper is structured as follows: In Section~\ref{sec:challenge}, we introduce the datasets of MSP-Podcast and the task, while section~\ref{sec:relatedwork} position our study in regards with some related works. Section~\ref{sec:overview_approach} presents the general architecture of our system. The components of the sub-systems are described in Section~\ref{sec:components}. The sub-systems themselves are presented in Section~\ref{sec:sub_systems}, as well as the fusion approach in Section~\ref{sec:fusions}. Finally, in Section~\ref{sec:results}, we present the results and discuss them.

\section{MSP-Podcast SER Challenge 2024}
\label{sec:challenge}

MSP-Podcast\cite{Lotfian_2019_3} is a large naturalistic speech emotion corpus featuring speech segments sourced from an audio-sharing website. The corpus is annotated for both categorical emotion recognition and emotional attribute prediction. The training set consists of 68,119 speaking turns and the development set contains 19,815 speaking segments. The test set comprises 2,347 unique segments from 187 speakers, with the labels not publicly disclosed. The selection of segments for the test set was carefully curated to ensure a balanced representation across primary categorical emotions.

Two tasks were proposed, we are only participating in the first task. The first task involves categorical classification within eight specified emotional states: Anger (A), Happiness (H), Sadness (S), Fear (F), Surprise (U), Contempt (C), Disgust (D), and Neutral (N). The test set for the challenge has a balanced distribution across the emotional categories. Performance evaluation and model ranking on the leaderboard are based on the Macro-F1 score. The Macro-F1 score is calculated by first computing the F1 score separately for each class, which is the harmonic mean of precision and recall for that class, and then taking the average of these F1 scores.


\section{Related Work}
\label{sec:relatedwork}

\subsection{Speech Emotion Recognition}
\gls{SER} involves a two-step process: feature extraction and emotion classification.  
Early \gls{SER} research focused on handcrafted features like pitch, energy, and Mel-frequency cepstral coefficients (MFCCs) along with traditional machine learning methods, including Markov models, Gaussian mixture models, and support vector machines for classification~\cite{schuller09_interspeech, Schuller2013}.  
More recently, neural-based models have started to replace traditional machine learning approaches~\cite{schuller09_interspeech}\cite{Schuller2013}.  Convolutional Neural Networks (CNNs) and Recurrent Neural Networks (RNNs) have demonstrated improved performance in emotion recognition tasks~\cite{li19n_interspeech, Etienne2018}.  
Additionally, transfer learning, particularly self-supervised pretraining, has gained traction in \gls{SER}. Models like Wav2Vec2 2.0~\cite{baevski2020wav2vec}, WavLM~\cite{chen2022wavlm} and HuBERT~\cite{lakhotia2021generative} have achieved state-of-the-art results in this domain~\cite{macary2020use, wang2022finetuned}. \newline

\subsection{Text Emotion Recognition}
Text Emotion Detection and Recognition (TEDR) has significantly evolved over the past few years, transitioning from traditional machine learning approaches to deep learning models.  
Initial methods primarily utilized classifiers such as \gls{SVM} and Maximum Entropy (MaxEnt) classifiers, later approaches increasingly relied on deep learning models in combination with different word embedding methods.  
For example, an emotion detection model that combines Long Short-Term Memory (LSTM) networks with Convolutional Neural Networks (CNN) was introduced~\cite{polignano2019comparison}.  
This model integrates various word embeddings, including GloVe~\cite{pennington2014glove} and FastText~\cite{joulin2016fasttext}, to capture semantic nuances more effectively.    
More recently, approaches based on transformer pre-trained language models~\cite{adoma2020comparative} have begun to emerge, offering remarkable breakthroughs in text emotion detection.  
In~\cite{adoma2020comparative}, the authors conducted a comprehensive comparison of models including BERT~\cite{devlin2018bert}, RoBERTa~\cite{liu2019roberta}, DistilBERT~\cite{sanh2019distilbert}, and XLNet~\cite{yang2019xlnet}, with RoBERTa emerging as the model achieving the best performance.

\section{Overview of the approach}
\label{sec:overview_approach}

The system was developed as a two-level architecture. Given a speech segment, the first level extracts outputs from categorical emotion recognition based on various sub-systems. The outputs of theses sub-systems are fed to a \gls{SVM}. Five different set of sub-systems are used.

\begin{figure}[!ht]
    \centering
    \includegraphics[width=0.15\textwidth]{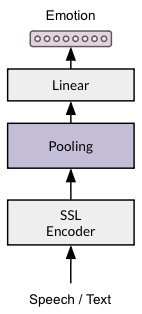}
\caption{Illustration of our speech emotion recognition system.}
\label{fig:ssl_system}
\end{figure}

We formulate the sub-system as a mapping from the continuous speech domain into the discrete domain of categorical labels of emotion. As depicted in Figure~\ref{fig:ssl_system}, to achieve this, we first use an encoder (speech and/or text encoder). These encoders have been trained on unlabeled data and are capable of extracting highly robust feature representations. Following the encoder stage, we employ a pooling strategy to aggregate these features over time, ensuring a fixed-length representation regardless of the original speech duration. This representation then feeds into a classifier layer, which serves as the final component of our architecture. This layer is responsible for mapping the aggregated features to the desired categorical labels of emotion, thus completing the process of transforming continuous speech into discrete emotion predictions.


\section{Sub-System components}
\label{sec:components}

In this section, we describe the various components used in the sub-systems. In Section~\ref{sec:encoder}, we describe the various speech and text encoders used in our study. Following that, Section~\ref{sec:pooling} is dedicated to detailing the pooling techniques employed. The discussion continues in Section~\ref{sec:classifier}, where we delve into the classifier used. Finally, Section~\ref{sec:data_augmentation} covers the aspects of speech audio data augmentation.

\subsection{Encoder}
\label{sec:encoder}

In the realm of speech and text processing, the choice of encoders is essential for provide powerful deep feature learning. Indeed, these encoders are trained on large unannotated datasets, enabling them to learn rich, complex patterns without the need for manually labeled data. 

For speech encoder, we leverage the capabilities of WavLM~\footnote{https://huggingface.co/microsoft/wavlm-large}, a state-of-the-art \gls{SSL} speech model designed to discover speech representation that encode pseudo-phonetic or phonotactic information. WavLM is distinguished by its robustness in handling a wide range of audio scenarios, including noisy environments. Additionally, we also used HuBERT~\footnote{https://huggingface.co/facebook/hubert-large-ll60k}, another leading speech encoder model. For both model, features are extracted, at the acoustic frame-level i.e. for short speech segments of 20 ms duration. 

For text encoder, we employ RoBERTa~\footnote{https://huggingface.co/FacebookAI/roberta-base}, which is builds on the BERT architecture but optimizes its pre-training process for more efficient learning and better performance across a variety of \gls{NLP} tasks. Features are extracted for every words.

\subsection{Pooling}
\label{sec:pooling}

Given that speech segments vary in length, we use pooling to aggregate the features given by the encoder across time, ensuring a fixed-length representation regardless of the original speech duration. Two different types of pooling have been used:

\begin{itemize}

\item \textbf{mean-pooling} achieves this by averaging the feature values given by the encoder over the time dimension, which effectively summarizes the overall characteristics of the speech signal into a single unified representation.

\item  \textbf{attention-pooling} unlike mean-pooling leverages a weighted average, where the weights are learned through the model~\cite{safari2020selfattention}. This allows the model to focus on more relevant parts of the speech signal, potentially capturing nuanced dynamics that mean pooling might overlook.

\end{itemize}

\subsection{Classifier}
\label{sec:classifier}

For the classifier component, we assume that the \gls{SSL} encoders have successfully captured all necessary information for predicting the targeted categorical emotion label.  
We choose a simple yet efficient approach by integrating a linear layer to function as the classifier.
This linear classifier takes the pooled feature vectors and assigns them to the a given emotion label.

\subsection{Data Augmentation}

\label{sec:data_augmentation}
The MSP-Podcast corpus includes, for each segment, the emotion label, text transcription, speaker ID, and gender. However, the test set is provided as speech only, without any annotations.  
To utilize a text encoder, we needed to automatically generate transcriptions for the test set.  
For this purpose, we employed the Whisper~\cite{radford2022robust} speech recognition model.  
Additionally, to ensure consistency and avoid any discrepancies between provided transcriptions and Whisper-generated transcriptions, we computed transcriptions for the entire dataset.  \\

\noindent We observed a significant discrepancies in the distribution of emotion classes within both the Training and Development sets. Additionally, a considerable number of samples are labeled as "X," indicating that no consensus was reached for these samples. To minimize the mismatch, we decided to automatically recompute the consensus for samples labeled as "X". For each sample, we have access to the labels provided by each annotator and the consensus. The consensus is determined by identifying the most frequently associated label. In scenarios where no single label predominates due to an even distribution of votes among the labels, the label "X" is assigned to indicate the lack of a clear consensus.  \\
Our method for recomputing the consensus is detailed as follows: First, we calculate a score for each evaluator by determining the ratio of their annotations that match the consensus to their total number of annotations. Then, we discard all annotations from evaluators whose score falls below a predefined threshold $K$. Finally, we recompute the consensus for the whole dataset.  
In scenarios where there is a tie between a specific label and the Neutral ("N") label, we opt to drop the label "N". Samples with newly attributed labels from the Training and Development sets were added to the training set.

\begin{table*}[h]
 \centering
 \caption{Performance comparison of individual sub-systems and their fusion across all emotion classes, with overall accuracy and F1-Macro scores.}
 \label{tab:results}
 \begin{tabular}{c|c|c|c|c|c||c}
    \toprule
      & Sub-System A & Sub-System B & Sub-System C & Sub-System D &  Sub-System E & Fusion  \\
    \midrule
    Anger (A)        & \numprint{0.5462} & \numprint{0.5521} &  \numprint{0.5242} & \numprint{0.4874} & \numprint{0.4902} & \numprint{0.4978} \\
    Contempt (C)        & \numprint{0.1412} & \numprint{0.1040} &  \numprint{0.1642} & \numprint{0.1983} & \numprint{0.2325} & \numprint{0.2712} \\    
    Disgust (D)        & \numprint{0.1178} & \numprint{0.1657} &  \numprint{0.1643} & \numprint{0.1910} & \numprint{0.1605} & \numprint{0.1300} \\    
    Fear (F)         & \numprint{0.0331} & \numprint{0.0259} &  \numprint{0.0553} & \numprint{0.0642} & \numprint{0.0268} & \numprint{0.0272} \\    
    Happiness (H)         & \numprint{0.5886} & \numprint{0.5772} &  \numprint{0.5507} & \numprint{0.5910} & \numprint{0.6135} & \numprint{0.6236} \\    
    Neutral (N)         & \numprint{0.6079} & \numprint{0.6036} &  \numprint{0.5949} & \numprint{0.5358} & \numprint{0.6062} & \numprint{0.6321} \\    
    Sadness (S)         & \numprint{0.3147} & \numprint{0.3600} &  \numprint{0.3181} & \numprint{0.3810} & \numprint{0.3855} & \numprint{0.3948} \\    
    Surprise (U)         & \numprint{0.2020} & \numprint{0.1863} &  \numprint{0.1742} & \numprint{0.2108} & \numprint{0.1694} & \numprint{0.2220} \\        
    \midrule
    Accuracy &  \numprint{0.5087} & \numprint{0.5083} & \numprint{0.4843} & \numprint{0.4659} & \numprint{0.5046} & \numprint{0.5217} \\
    F1-Macro &  \numprint{0.3189} & \numprint{0.3218} & \numprint{0.3182} & \numprint{0.3324} & \numprint{0.3356} & \numprint{0.3498} \\
    \bottomrule
 \end{tabular}
\end{table*}

\section{Sub-Systems}
\label{sec:sub_systems}

In this section, we provide a detailed description of the five distinct sub-systems employed in the fusion process.

\subsection{Sub-System A : WavLM}


Thi system is based on a WavLM, a mean pooling strategy and the outoput is a softmax loss function. In order to optimize this architecture, we employ the Adam optimizer. We fine-tune the pre-trained WavLM model during the training phase. This fine-tuning allows the WavLM model to adjust its parameters specifically towards recognizing emotions in speech, leveraging the rich, pre-learned representations and tailoring them to our domain of interest. We set the mini-batch size to 16 and 10 steps. And no data-augmentation is done on speech segment.

\subsection{Sub-System B : Jeffreys Loss}

This system is identical to System A; however, we propose in this system to replace the softmax loss function by Jeffreys loss function. The Jeffreys Loss is given in the Equation~\ref{eq:eq_final_3}. Incorporating this divergence into the cross-entropy loss function enables the maximization of the target value of the output distribution while smoothing the non-target values.

\begin{equation}
\mathcal{L}=-\log\left(  p_{k}\right)  -\alpha\frac{\sum_{i\neq k}\log p_{i}}{K-1} +\beta\frac{\sum_{i\neq k}p_{i}\log p_{i}}{1-p_{k}}
\label{eq:eq_final_3}
\end{equation}

\subsection{Sub-System C : Joint Wav2vec2-WavLM}

\begin{figure}[!ht]
    \centering
    \includegraphics[width=0.33\textwidth]{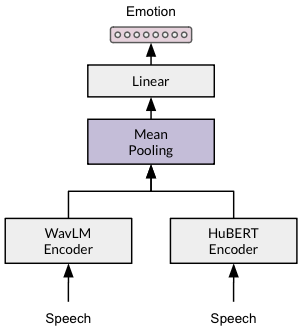}
\caption{Illustration of the dual speech encoder emotion recognition system.}
\label{fig:dual_speech_encoder}
\end{figure}

This system is based on a joint \gls{SSL} audio strategy as depicted in Figure~\ref{fig:dual_speech_encoder}, wherein the upstream model is performed by jointly training a WavLM and Hubert \gls{SSL} model. The input speech is fed into both \gls{SSL} models. Fine-tuning of the upstream model is achieved throught simultaneous training alongside a straightforward network that implements mean pooling, leading to a Linear Classifier, as illustrated in Figure 2.

\subsection{Sub-System D : WavLM and RoBERTa}
\label{sys_a}
\begin{figure}[!ht]
    \centering
    \includegraphics[width=0.33\textwidth]{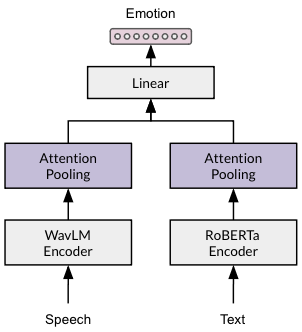}
\caption{Illustration of our joint speech and text emotion recognition system.}
\label{fig:joint_system}
\end{figure}

The following describes the \gls{SER} model depicted in Figure~\ref{fig:joint_system}.  
The WavLM encoder is divided into two parts: the CNN feature extractor and the trasnformer-based encoder.  
We chose to freeze the CNN feature extractor part, fixing all the parameters of these CNN layers and only fine-tune the parameters of transformer blocks.  
This method of partial fine-tuning acts as a strategy for domain adaptation. It is designed to maintain the integral feature extraction capabilities of the lower layers, thereby enabling the model to adjust to new tasks efficiently without compromising its overall performance.  
For the text encoder, we opt to fine-tune all parameters of the RoBERTa model.
During the fine-tuning process, we apply three different schedulers to adjust the fine-tuning learning rates of the WavLM and RoBERTa encoders, as well as the learning rate of the classifier model. 
Each scheduler utilizes the Adam Optimizer in conjunction with the NewBob technique, which anneals the learning rates based on the performance during the validation stage.
The fine-tuning learning rates for both the WavLM and RoBERTa encoders are set to $10^{-5}$, while the learning rate for the classifier model is set to $10^{-4}$.
This model is trained using negative log-likelihood loss (NLL).

\subsection{Sub-System E : Data Augmentation}
The system E is identical to the system D described in section~\ref{sys_a}.
However, this system incorporates the data augmentation technique outlined in Section~\ref{sec:data_augmentation}, setting $K=50\%$. 
This adjustment aims to mitigate the notable imbalances observed in the distribution of emotion classes.

\begin{table}[htbp]
\centering
\caption{Distribution of emotion classes before and after data augmentation.}
\label{tab:distribution_aug}
\begin{tabular}{@{}c|c|c@{}}
\toprule
Emotion Class & Train              & Train Augmented              \\ \midrule
N             & 25106              &  25106                       \\
H             & 13440              &  13440                       \\
S             & 3882               &  6067                        \\
A             & 3053               &  4753                        \\
U             & 2897               &  4328                        \\
C             & 2443               &  2897                        \\
D             & 1426               &  2352                        \\
F             & 1139               &  1681                        \\ \midrule \midrule
Total         & 53386              &  60624                       \\ \bottomrule
\end{tabular}
\end{table}

\section{Fusion systems}
\label{sec:fusions}

The fusion process integrates the outputs from all five sub-systems (A, B, C, D, E) by concatenating them to form a single feature vector. 
This concatenated vector serves as the input to an SVM classifier, which is trained to predict the emotion class. 
This fusion approach leverages the diverse strengths of each sub-system, aiming to enhance the overall performance and robustness of emotion classification.

\section{Experiments}
\label{sec:results}

In this section, we describe our experimental setup, analyze results from individual sub-systems and their combined fusion.

\subsection{Experimental setup}
To conduct our experiments, we employed the SpeechBrain toolkit \cite{speechbrain}, which is built on PyTorch and is specifically designed for speech-processing tasks. Additionally, we utilized the Hugging Face versions of the WavLM, HuBERT, and RoBERTa models. The source code is available on GitHub\footnote{github.com/Chaanks/MSP-Podcast-SER-Challenge-2024}.
\subsection{Results}

\noindent \textbf{Impact of the Speech encoder:} A first experiments aimed at identifying the most effective Speech encoder. Various speech encoders, including Wav2Vec, Hubert and WavLM were evaluated. These experiments were carried out on sub-system D. Table~\ref{tab:results_speech_encoder}, give the results of the experiments, from which we observe that the best results are obtained with the WavLM.

\vspace{10pt}

\noindent \textbf{Impact of the Text encoder:} An additional experiment was conducted using text as input. 
This experiment, as shown in Table~\ref{tab:results_speech_encoder}, reveals that the performance of the RoBERTa text encoder, achieving a Macro-F1 score of $0.27$, is not too far from the best speech encoder, which has a Macro-F1 score of $0.32$.
This result encourages the exploration of a joint speech and text emotion recognition system.

\begin{table}[htbp]
 \centering
 \caption{Performance Comparison of Speech and Text Encoders. The table shows Micro-F1 and Macro-F1 scores for speech encoders (Wav2Vec2, WavLM, HuBERT) and a text encoder (RoBERTa).}
 \label{tab:results_speech_encoder}
 \begin{tabular}{c|c|c}
    \toprule
    Model  & Micro-F1 $\uparrow$ & Macro-F1 $\uparrow$ \\
    \midrule
    Wav2Vec2 Large & 0.45 & 0.29 \\
    \midrule
    WavLM  Large & 0.51 & 0.32 \\
    \midrule
    HuBERT Large & 0.50 & 0.31 \\
    \midrule
    \midrule
    RoBERTa Base & 0.44 & 0.27 \\
    \bottomrule
 \end{tabular}

\end{table}

\vspace{10pt}
\noindent \textbf{Impact of the Fusion:} Table~\ref{tab:results} shows the results achieved by the different sub-systems. We observe that all the sub-systems achievd F1-Macro scores between $0.32$ and $0.34$. The best sub-system is sub-system E. And we observe that fusion system led to an improvement, achieving an F1-Macro score of 0.35\%.

\begin{figure}[!ht]
    \centering
    \includegraphics[width=0.40\textwidth]{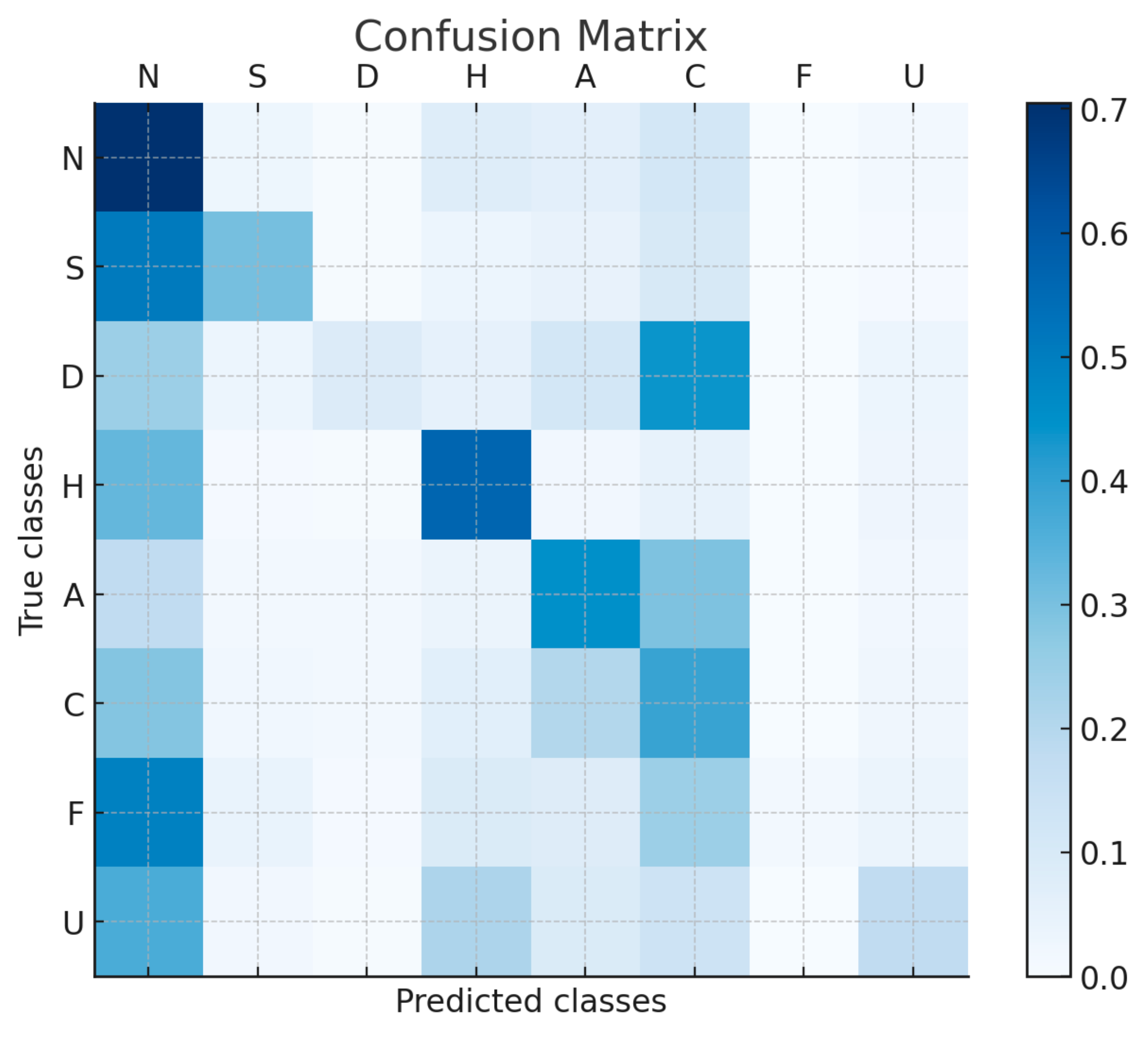}
\caption{The confusion matrix provided by the fusion system.}
\label{fig:confusion_matrix}
\end{figure}

\vspace{10pt}
\noindent \textbf{Confusion matrix:} Figure~\ref{fig:confusion_matrix} shows the confusion matrix obtained from fusion system.
Regarding the diagonal elements, we can observed that the fusion model is particularly effective at correctly predicting Neutral (N) and Happiness (H) classes. However, the fusion system struggles more with accurately predicting the Disgust (D) and Fear (F) classes.
As for the off-diagonal elements, we can noted that there is a tendency for all classes to be misclassified as Neutral (N) and Disgust (D) misclassified as Contempt (C).

\section{Conclusion}
This paper describes LIA's participation in the MSP-Podcast \gls{SER} Challenge. Our approach involves developing sub-systems, each of which functions as an emotion classifier. These sub-systems model speech segments using different components to provide varied perspectives. For the final fusion step, we concatenate the outputs of the sub-systems and train a \gls{SVM} for fusion. The fusion system achieved an F1-Macro score of 0.35\% on the development set.

\section{Acknowledgements}

This work was granted access to the HPC resources of IDRIS under the allocation AD011013257R2 made by GENCI.

\newpage

\bibliographystyle{IEEEbib}
\bibliography{Odyssey2024_BibEntries}

%

\end{document}